\documentclass[twocolumn]{article}
\usepackage{amssymb}
\usepackage{moreverb}
\usepackage[cmex10]{amsmath}
\usepackage{graphicx}
\usepackage{makecell}
\usepackage{hyperref}
\usepackage{titling}
\newcommand{\subtitle}[1]{%
  \posttitle{%
    \par\end{center}
    \begin{center}\large#1\end{center}
    \vskip0.5em}%
}

\begin{document}

\title{P-Tree Programming}
\subtitle{(Working Paper)}
\author{Christian Oesch\\
\small University of Basel, Switzerland
}

\maketitle

\begin{abstract}
We propose a novel method for automatic program synthesis. P-Tree Programming represents the program search space through a single probabilistic prototype tree. From this prototype tree we form program instances which we evaluate on a given problem. The error values from the evaluations are propagated through the prototype tree. We use them to update the probability distributions that determine the symbol choices of further instances. The iterative method is applied to several symbolic regression benchmarks from the literature. It outperforms standard Genetic Programming to a large extend. Furthermore, it relies on a concise set of parameters which are held constant for all problems. The algorithm can be employed for most of the typical computational intelligence tasks such as classification, automatic program induction, and symbolic regression.
\end{abstract}

\section{Introduction}
Automatic program synthesis has a long history in computational intelligence. Especially Genetic Programming (GP) \cite{koza1992genetic} has produced many human-competitive results in engineering, game playing, image recognition, mathematical algorithms, robotics, software repair, reverse engineering, and empirical model discovery \cite{koza2010human}.  GP evolves a population of solutions through an algorithm mimicking natural selection. By iteratively creating, selecting and combining candidate solutions, impressive results can be achieved. However, there are a few issues with the GP algorithm. First of all, after a few generations, it is very unlikely for new genetic material to survive, as the natural selection pressure eradicates candidates independent of the selection scheme rapidly \cite{back1994selective}. As a result, the population converges quickly and often gets stuck in a local optima. A second major drawback of GP is that only individuals have a fitness measure assigned to. Sub-branches or nodes of the program tree do not exhibit a figure of merit.  A third disadvantage of GP is that when an individual is excluded from the population, all its information is lost. When a individual could have evolved into a promising candidate, it is unlikely to occur again.

We propose P-Tree Programming (PTP), a new method for automatic program synthesis based on a single prototype tree. Instead of keeping a population of individuals, we require only one fully expanded tree. The tree represents the probabilities of drawing a certain program instance. Only the error values of already evaluated choices of the prototype tree have to be kept in memory. This makes the endeavor computationally feasible. The method works similar to Probabilistic Incremental Program Evolution \cite{salustowicz1997probabilistic}. However, our method differs on how the error values determine the probability distributions of the nodes. A choice's error value is the minimum error value of all its sub-branches. This allows us to keep all information from the evaluated solutions and prevents the algorithm from pre-mature convergence on a local optima. Additionally, the information encoded in the branches of the tree is exploited naturally by the algorithm.

In this study we describe PTP and apply it to a set of symbolic regression benchmarks. In symbolic regressions we search for a symbolic expression that describes the input-output relationship of a given data set the best. We chose these symbolic regression benchmarks because they are easily understood. Furthermore, there is a broad literature comparing a variety of methods to those particular problems

In the next Section we describe the basic method in detail. We also discuss some simple extensions: the discounting of the error values and the creation of constants in the framework. Section \ref{sec:experiments} describes the problem set as well as the implementation details of this study. Additionally, we provide a Cython implementation of the algorithm. Section \ref{sec:results} shows the results from the experiments and discusses some of the implications of the method. Finally, we conclude in Section \ref{sec:conclusion}.

\section{Method}\label{sec:method}
\subsection{Representation}
\begin{figure}
\center
\includegraphics{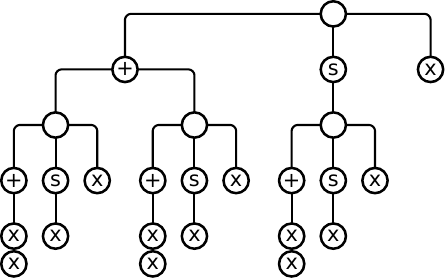}
\caption{The tree of possible programs for a simple function set with non-terminals \texttt{\{+, s\}} and terminal \texttt{\{x\}}. A maximum program tree depth of three is enforced. Empty circles represent a node of the tree and circles with symbols represent a choice. Note that \texttt{+} links to two nodes.}
\label{fig:ptree1}
\end{figure}
We are searching for a symbolic expression that satisfies a given criteria. We call this expression a program. Given a certain function set, all possible programs from the set can be represented in one tree structure. In accordance with \cite{salustowicz1997probabilistic}, we call this the prototype tree. The prototype tree is made of nodes which link to further nodes containing the arguments of the function choices made by the nodes. Note that the prototype tree is different to a tree in GP, as the nodes just represent an argument to a function and not the choice of the symbol itself. Each node has a set of symbols to choose from. To each of these nodes a (multiple) specific node(s) is (are) linked if the symbol is a non-terminal. Symbols which require a further argument (e.g. \texttt{+, sin( ), log( ), \dots}) are called non-terminals. Symbols which do not require a further argument (e.g. a variable \texttt{x}) are called terminals.

As an example consider a function set made up of the symbols \texttt{\{+, s, x\}}. The symbol \texttt{+} represents addition, \texttt{s} is the sine and \texttt{x} is an input variable to the program. A visual representation of the tree of all possible programs with a maximum depth of three can be found in Figure \ref{fig:ptree1}. Empty circles represent a node of the tree and circles with symbols represent a symbol choice. The root node chooses from the full set of choices \texttt{\{+, s, x\}}. Depending on the arity of a function, the choices require a different amount of linked nodes and choices. The symbol \texttt{+} requires two further nodes, \texttt{s} requires one further node and \texttt{x} requires no further node.

We build a program instance from the prototype tree by choosing an initial symbol from the set of all symbols. This means, we have 3 possibilities. If \texttt{x} is chosen, the program is complete, as all of the tree's leafs contain a terminal. If the symbol \texttt{s} (\texttt{+}) is chosen, one (two) further symbol(s) is (are) required. The process is repeated until the terminal \texttt{x} occurs. Without a limitation, this tree can grow to infinite depth. It is common to impose a maximum depth, at which point only terminal symbols are allowed. Other mechanics to limit the size of instances are possible, as seen in Section \ref{subsec:discounting}

\subsection{Search}
In supervised learning we want to create a mapping from input arguments to the output, that best satisfies our requirements. In GP, this is handled by randomly creating multiple realizations of a program, so called individuals, to create a population. Each individual in the population is then evaluated. Well-performing solutions are combined and mutated to create ever better versions. To save computing time, bad solutions die and leave the population.

In PTP, we search for the best choices of a node given its  position in the tree. We keep the error of the nodes' choices in memory and only ever have one realization of the program. The probabilities for drawing a particular instance are influenced by the performance of previously evaluated instances. 

\begin{figure}
\center
\includegraphics{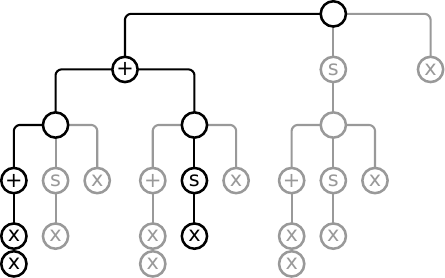}
\caption{One realization from the prototype tree from Figure \ref{fig:ptree1}. It evaluates to \texttt{x+(x+sin(x))}.}
\label{fig:ptree2}
\end{figure}

Initially, each node in the prototype tree has a probability distribution that weights each choice with equal probability 
$$p_c=\frac{1}{N},$$ 
where $N$ is the number of symbols in the function set. Note that this uniform distribution does not need to be kept in memory for all nodes of the prototype tree. Only for choices that have been evaluated, the error values are stored. This means that while the prototype tree represents the full search space, we do not have to keep it in memory. 

The first realization of a program is then drawn from the prototype tree. We start from the root node and draw a uniformly distributed random variable which decides the root node's choice. We then step down one level and decide the choices of the linked nodes in the same manner. This process is repeated until the program is complete. A program is complete when all leafs of the tree contain a terminal symbol. Figure \ref{fig:ptree2} shows a possible realization (\texttt{((x+x)+sin(x)}) of the example prototype tree from the last subsection.

This first program instance is evaluated on the training-set and an error (or to use the GP terminology, a fitness score) is calculated. For symbolic regression problems this can be for example the mean squared error (MSE). Subsequently, the error is propagated from the root node through the tree to each of the nodes which make up the current instance. For each choice contained in the instance, the error value replaces the current value of the node if it is smaller than the currently stored error. This means that a choice's value in the prototype tree is determined by the best of its sub-branches.  An example of this can be seen in Figure \ref{fig:ptree3}. The minimum error in this example is 2.28. It is formed by the program \texttt{s(x+x)}. The value 2.28 now represents the choices for \texttt{s} in the root node, as well as the choice for \texttt{+} in the node on the first level. Note that this is one of the main differences to {\cite{salustowicz1997probabilistic}. We value a branch by the best of its results. Consider another example in Figure \ref{fig:ptree3}. If the expression \texttt{s(s(x))} has a score lower than 2.28, it will replace the error assigned to the choice \texttt{s} in the root node. Otherwise \texttt{s(x+x)} will remain the best choice and the value of \texttt{s} remains at 2.28. 

This mechanism creates a path through the tree with minimum error values. If one follows this path, it creates the best expression evaluated so far. In fact, if applied multiple times, it builds various paths through the search space.  We suggest this structure can be used to explore the prototype tree in an efficient way. It is easy to balance exploration and exploitation by deciding how often we stay on the best path.

\begin{figure}
\center
\includegraphics{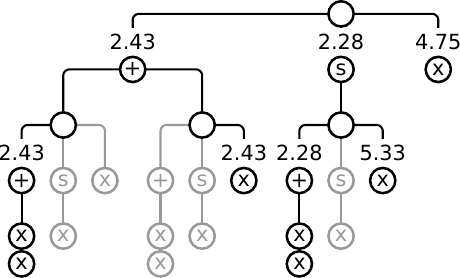}
\caption{The black branches have been evaluated and the error values have been assigned to the choices for the evaluated nodes. Note how the minimum (2.28) of the values of the sub-branches for the choice of \texttt{s} in the root node is used to represent its value.}
\label{fig:ptree3}
\end{figure}

Further realizations of the tree are drawn and the process is repeated. Once all choices of a node have been evaluated at least once, a new probability distribution is calculated for the node. The probability for a choice is given by a power-law based on the error values of the choices:
$$p_c = \frac{r_c^{-k}}{\sum_{i}^{N} r_i^{-k}}$$
where $r_c$ is the rank of a choice starting from $1$ for the smallest error to $N$ for the largest error. The parameter $k$ (usually between 3.5 to 4.5) balances the exploitation versus exploration behavior of the search process. Basing the probabilities on the ranks instead of the actual error size has proven to be very effective for problems with large differences in the errors. In the following, whenever a node's error values are updated, the probability distribution is recalculated.

Again, as above, the probabilities build paths through the search space. Following the choices with the highest probability will lead to the best solution found so far. Branching of this path are a multitude of paths with high probability assessments. This builds a probability structure similar to a propensity map. However, the probabilities of choices are dependent on choices higher up in the tree. The process of drawing a new instance, evaluating it, assigning the error to the corresponding nodes, and calculating a new probability distribution is repeated.  The algorithm is stopped once a termination criterion, such as a minimum error value or a maximum number of iterations, is met.

\subsection{Discounting factors}\label{subsec:discounting}
There are several ways of improving the behavior of the algorithm. To combat bloat (the growth of the expression in size) of the representations, we discount the error values by the depth of a choice's branch. The error value of the choice $\epsilon$ is discounted by:
$$\epsilon_\text{choice} = \epsilon(1+\delta_d)^{(d_\text{node} - d_\text{branch})}$$
where $d_\text{node}$ is the depth of the node, $d_\text{branch}$ the depth of the branch's leaf node, and $\delta_d$ is a discounting factor, typically in the range of 0.0001 to 0.001. The discounting based on branch depth will guide the algorithm to choices that lead to instances with fewer nodes.

The algorithm can get stuck in a local optima for many iterations if a solution is hard to find. We can penalize choices that have been evaluated many times without finding a better solution in its sub-branches. If no better solution has been found and the choice was part of the evaluated solution, the current best error of the choice $\epsilon_{\text{choice}, i}$ is discounted by:
$$\epsilon_{\text{choice}} = \epsilon(1+\delta_p)$$
where $\delta_p$ is a discounting factor. The discounting factor $\delta_p$ should be small, typically in the range of 0.0001 to 0.01. For simple problems this is not needed. However, in harder problems this can improve the performance of the algorithm.   

\begin{figure}
\center
\includegraphics{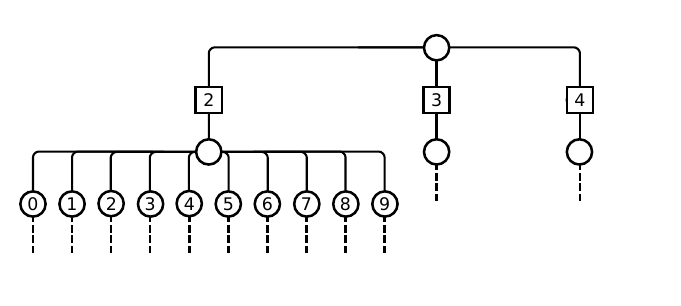}
\caption{An example of a constant branch. The square nodes determine the position of the floating point (i.e. $c = \gamma\times 10^{-m}$, where $m$ is the value of the float node) and the round nodes are used for digit concatenation. Only part of the tree is drawn. For a digit nodes depth of 3,  the example branch can represent floats from 0.01 to 9.99.}
\label{fig:ptree_const}
\end{figure}

\subsection{Constant creation}\label{sec:const}
To create constants we introduce two special nodes. The first type of node, the float node, determines the position of the floating point. A chain of the second type, the digit nodes, creates an integer with a fixed number of digits defined by the depth of the structure.

Figure \ref{fig:ptree_const} shows a branch that creates a constant. If a constant is chosen, a first node decides on the position of the floating point. In the example, the values 2, 3, or 4 are available. The value $m$ of this float node is used to multiply the integer $\gamma$, created by digit concatenation of the following nodes, to produce the final constant $c$:
$$c = \gamma \times 10^{-m}$$

The integer $\gamma$ is produced by concatenating the choices of the nodes following the float node. For example, in the first node, the value 3 is chosen, and in the next two nodes the value 1 is chosen. This results in the integer 311. If the choice of the float node was 4, the resulting constant $c$ is $311\times 10^{-4} = 0.0311$. This is only a suggestion to create constants. We suspect there are better performing ways to create constants. It would be straight forward to adapt the Levenberg-Marquardt algorithm \cite{levenberg1944method} to the system, as is applied to GP in \cite{kommenda2013nonlinear}.

\subsection{Implementation quirks}
Two simple adjustments to the basic algorithm have proven themselves useful for the implementation. First, if our prototype tree is very deep and we have evaluated many of its nodes, we might run into memory limitations. In this case, we can run a pruning pass where we remove certain nodes from the tree which are unlikely to contain the final solution. This step is not included in the experiments of the next section, as it was not needed for these rather small problems. The best way to do this is still subject to research.

Second, it can be useful to bias the probability of an unevaluated node, such that in a first evaluation (and only in the first) always a terminal symbol is chosen. Many nodes will only be evaluated once. After the very first evaluation, the probabilities for the symbols are again set to $1/N$ and the algorithm continues as usual. This bias can decrease the execution time of the algorithm as it creates fewer very deep solutions that are of no interest. In preliminary experiments, we have found that no notable decrease in the general performance of the algorithm occurs.

\section{Experiments}\label{sec:experiments}
\begin{table*}
\center
\caption{Problem set}
\small	
\begin{tabular}{llcc}
	\hline
name&target function & function set& training \\
& & & testing \\\hline

$\text{Nguyen-4}$&$ := x^6 + x^5 + x^4 + x^3 + x^2 + x$ &$+,-,*,/,\sin, \cos, \exp, \log, x, 1$&$U[-1,1, 20]$\\
&&&none\\
$\text{Nguyen-7}$&$ :=\log(x+1) + \log(x^2+1)$ &$+,-,*,/,\sin, \cos, \exp, \log, x, 1$ & $U[0,2, 20]$ \\
&&&none\\
$\text{Pagie-1}$&$ :=\frac{1}{1+x^{-4}}+\frac{1}{1+y^{-4}}$ &$+,-,*,/,pow,\sqrt, \log, ^{-1},x, y, c$& $E[-5,5,0.4]$\\
&&&$none$\\
$\text{Keijzer-6}$&$ :=\sum_i^x \frac{1}{i}$ &$+,*,\sin, \cos,\sqrt, \log,  ^{-1}, x, c$& $E[1, 50, 1]$\\ 
&&&$E[1, 120, 1]$\\
$\text{Korns-12}$&$ :=2-2.1\cos(9.8x) \sin(1.3w)$ & $+,-,*,/,\sin, \cos,\tan, \tanh,\sqrt, $&$U[-50,50,10000]$\\
&&$\exp,\log,^2, ^3, u,v,w,x,y,c$& $U[-50,50,10000]$\\
$\text{Vlad.-4}$&$ :=\frac{10}{5+\sum_{i=1}^5 (x_i -3)^2}$ &$+,-,*,/,pow,\sin, \cos, \sqrt, \exp, $& $U[0.05,6.05,1024]$\\
&&$ \log,^{-1}, u, v, w, x, y, c$&$U[-0.25,6.35,5000]$\\\hline

\end{tabular}
\label{tab:probset}
\end{table*}
To test the performance of the algorithm, we have chosen some well-known symbolic regression problems from the literature \cite{white2013better, uy2011semantically}. The target functions can be found in Table~\ref{tab:probset}. It lists the name of the problem, the target function that is to be discovered, the function set to the problem, as well as the training and testing domain of the variables. The notation $U[-1, 1, 20]$ means that the training (testing) set is created from a uniform distribution from -1 to +1, with 20 draws. The notation $E[1, 50, 1]$ on the other hand, describes a set of equidistant points in the range of 1 to 50 with a step size of 1. For all functions in the function set, the protected versions are used. For example, if a division by zero occurs, the division operator will return the value 1. The symbols $u, v, w, x, y$ represent the variables in the function sets. Note that Korns-12 has 5 variables in the function set but only 2 variables are required to solve the problem. The symbol $c$ represents a constant node as defined in Section \ref{sec:const}. For the calculation of the depth, we consider the constant a terminal symbol.

For the problems Nguyen-4, Nguyen-7, and Pagie-1 no testing set is defined in the original literature. Further, Nguyen-4 and Nguyen-7 only use a constant of 1.0 and do not allow for other constants. Thus, we include a constant of 1.0 in the function set and do not use the constant nodes. Especially the problems Korns-12, Vladsilasleva-4, and Pagie-1 are considered to be hard to solve with current symbolic regression methods.

\begin{table}
\center
\caption{Settings}
\small	
\begin{tabular}{lc}
\multicolumn{2}{c}{PTP}\\\hline
parameter & value\\\hline
iterations & 1000000\\
max depth & 15\\
max const. depth & 4\\
$m$ & $[1...6]$\\
$k$ & 4\\
$\delta_d$ & 0.001\\
$\delta_p$ & 0.00075\\
\multicolumn{2}{c}{GP}\\\hline
parameter & value\\\hline
population size& 5000\\
generations & 200\\
crossover prob. & 0.7\\
subtree mut. prob. & 0.1\\
hoist mut. prob. & 0.05\\
point mut. prob. & 0.1\\
parsimony coeff. & 0.01\\
tournament size & 20\\
\hline
\end{tabular}
\label{tab:settings}
\end{table}

To set the results into context, GP is applied to the same problems. The equivalent function set and an equivalent amount of function evaluations are provided. The settings for both methods can be found in Table \ref{tab:settings}. For PTP, the maximum tree depth is set to 15, the maximum depth of a constant branch is set to 3 and the floating point position is in the range of 1 to 6. The parameter $k$, defining the probability distribution, is set to 4 for all experiments. This has proven to be a good setting for a wide variety of problems. As an example, for the Keijzer-6 problem (nine symbols), this results in probabilities between 0.924 for the best to 1.41E-04 for the worst choice. The discounting factors are set to $\delta_d = 0.001$ and $\delta_p=0.00075$. For GP a large population size of 5000 individuals evolves for 200 generations. The crossover probability is set to 0.7, the subtree mutation probability to 0.1, the hoist mutation probability to 0.05 and the point mutation probability to 0.1. The parsimony coefficient, used to control bloat, is set to 0.01. For both PTP and GP the mean squared error (MSE) is used as a fitness measure.

The implementation of PTP\footnote{The code is available on Github \href{https://github.com/coesch/ptree/}{github.com/coesch/ptree}. It contains a range of benchmark functions. However, for brevity, only the results for the benchmarks listed in \cite{white2013better} and the well known polynomial problem are presented in this study.} is in Python 3.6 making use of the Cython extension \cite{behnel2011cython}. The gplearn \cite{Sthephens2016} library is used for the GP benchmarks. The experiments are run on an Intel Xeon (R) CPU E5-2609 machine at 8x2.4GHz with 32 GiB of memory running Ubuntu 17.04. To compare the two methods 100 independent runs have been performed.

\begin{table*}
\center
\caption{MSE}
\small	
\begin{tabular}{llcccccc}
	\hline
&&\multicolumn{2}{c}{PTP}&\multicolumn{2}{c}{PTP 1E5}&\multicolumn{2}{c}{GP}\\
name&&train&test&  train &test \\\hline
$\text{Nguyen-4}$&best&7.22E-34&none&1.17E-32&none&1.04E-04&none\\
 &median&2.49E-06&none&6.71E-05&none&1.15E-02&none\\
$\text{Nguyen-7}$&best&0.00E+00&none&0.00E+00&none&3.29E-04&none \\
 &median&1.95E-07&none&5.24E-06&none&1.33E-03&none\\
$\text{Pagie-1}$&best&6.72E-04&none&7.69E-03&none&3.46E-02&none\\
 &median&1.29E-02&none&2.19E-02&none&4.76E-02&none\\
$\text{Keijzer-6}$&best&1.48E-13&7.24E-14&6.16E-10&3.30E-10&8.00E-03&3.78E-03\\ 
 &median&1.22E-09&3.44E-09&1.47E-07&4.89E-07&8.02E-03&4.15E-03\\
$\text{Korns-12}$&best&1.08E+00&1.08E+00&1.08E+00&1.08E+00&1.08E+00&1.09E+00\\
 &median&1.11E+00&1.11E+00&1.11E+00&1.11E+00&1.11E+00&1.11E+00\\
$\text{Vlad.-4}$&best&3.00E-03&4.35E-03&1.44E-02&1.42E-02&3.38E-02&3.59E-02\\
 &median&1.22E-02&1.53E-02&1.44E-02&3.72E-02&3.93E-02&3.97E-02\\\hline
\end{tabular}
\label{tab:results}
\end{table*}

\section{Results}\label{sec:results}

\begin{figure*}
\center
\includegraphics{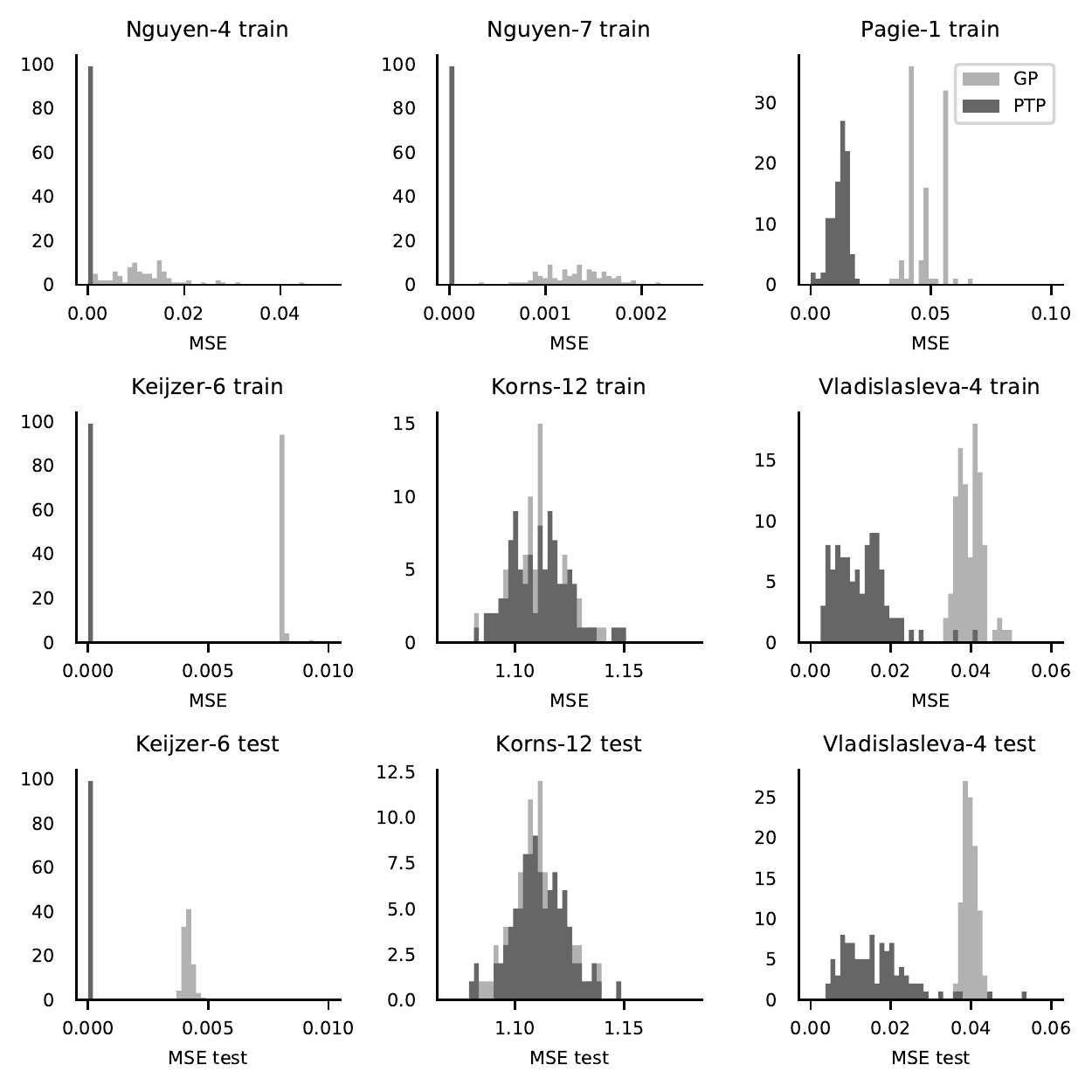}
\caption{Histograms of the end-of-training as well as the testing-set results (where applicable) for the problem set. The darker shaded values are from PTP. It clearly outperforms GP in most of the problem sets for training and testing.}
\label{fig:histos}
\end{figure*}

An overview of the results can be seen in Table \ref{tab:results}. Listed is the minimum as well as the median of the end-of-training and test results of the 100 runs for both PTP and GP. Additionally, the results of the PTP method after only 1E5 evaluations are shown as well. PTP outperforms GP in both measures for all the problems except for Korns-12. Especially for the problems Nguyen-4, Nguyen-7, and Keijzer-6 PTP's results are several magnitudes better. The Korns-12 problem seems to be very hard for both approaches. There exists a strong local optima for a constant close to 2.0. In all other problems, PTP already outperforms GP with only 10\% of the function evaluations.

Histograms of the MSE for the end-of-training and test MSE can be found in Figure \ref{fig:histos}. The top two rows show the end-of-training results. The bottom row shows the test results for the problem sets which include a testing sample. In some problems, even the worst run of PTP is still better than the best GP run.

In our implementation, execution of the 100 runs, utilizing 7 threads to perform some runs simultaneously, took the following amount of time: Nguyen-4: 1162.9s, Nguyen-7: 1781.0s, Pagie-1: 3036.7s, Keijzer-6: 1050.1s, Korns-12: 3242.2s, Vladislasleva-4: 4879.6s. A speed comparison to GP with gplearn is unfair, as gplearn is not compiled with Cython and requires much more execution time.

\begin{figure*}
\center
\includegraphics{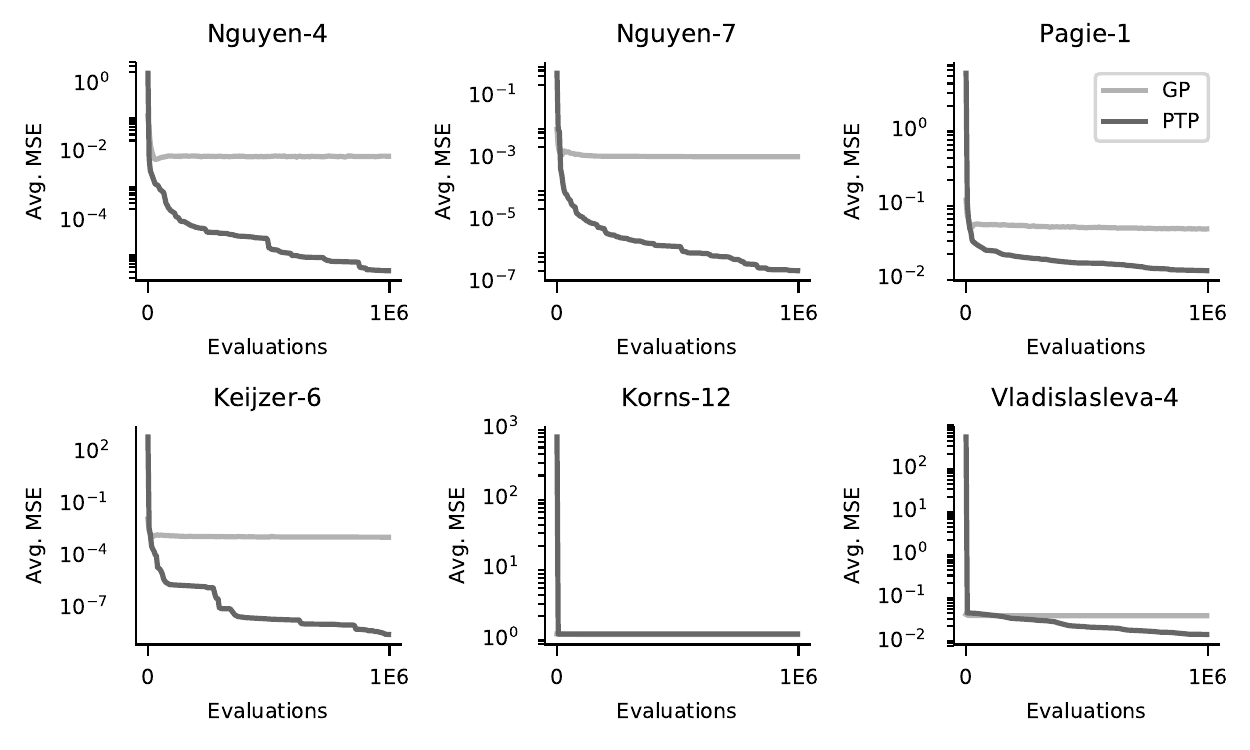}
\caption{Compares the evolution of the MSE on a log scale for PTP and GP. When GP has long stopped improving, PTP still gets better with more function evaluations.}
\label{fig:evol}
\end{figure*}
Figure \ref{fig:evol} shows the evolution of the MSE for both methods. While GP gets stuck in local optima, PTP continues to improve its solutions even up to 1'000'000 training set evaluations. More evaluations would probably lead to even better results in the benchmark problems. This could be, because PTP never excludes a solution completely, but only assigns a very low probability to it.

\section{Conclusion}\label{sec:conclusion}
We have presented P-Tree Programming, a novel method for automatic program synthesis. The method does not rely on a population of individuals, but only uses a prototype tree representing probabilities of node choices. From this prototype tree, instances are drawn based on probability distributions defined by previously evaluated candidate solutions. Further, we have shown simple extensions to the algorithm as well as how to create constants. However, these extensions are only examples. No claim is made that there is no better way to implement them within the method. 

To test PTP, we have applied it to a set of symbolic regression benchmark problems from the literature. The approach outperforms GP in most of the test problems. In only one of the benchmark problems, the algorithm did not improve on the performance of GP. This is where further extensions to the algorithm might be useful. The execution time of the algorithm should be similar to a very efficient implantation of GP, keeping function evaluations equal.

These initial results are very promising and further development of the algorithm should offer exiting new avenues of research. Many details of the approach still have to be fine tuned. The derivation of the probabilities, the way constants are evolved, and how the search space is explored are still subject to investigation. To improve the execution time, it is possible to sample only a part of the training set in each iteration. The algorithm could also be implemented in a non-recursive fashion on a GPU, promising further performance gains from heavy parallelization.
\bibliographystyle{plain}
\bibliography{literature}{}

\begin{thebibliography}{10}

\bibitem{back1994selective}
Thomas B{\"a}ck.
\newblock Selective pressure in evolutionary algorithms: A characterization of
  selection mechanisms.
\newblock In {\em Evolutionary Computation, 1994. IEEE World Congress on
  Computational Intelligence., Proceedings of the First IEEE Conference on},
  pages 57--62. IEEE, 1994.

\bibitem{behnel2011cython}
Stefan Behnel, Robert Bradshaw, Craig Citro, Lisandro Dalcin, Dag~Sverre
  Seljebotn, and Kurt Smith.
\newblock Cython: The best of both worlds.
\newblock {\em Computing in Science \& Engineering}, 13(2):31--39, 2011.

\bibitem{kommenda2013nonlinear}
Michael Kommenda, Michael Affenzeller, Gabriel Kronberger, and Stephan~M
  Winkler.
\newblock Nonlinear least squares optimization of constants in symbolic
  regression.
\newblock In {\em Revised Selected Papers of the 14th International Conference
  on Computer Aided Systems Theory-EUROCAST 2013-Volume 8111}, pages 420--427.
  Springer-Verlag New York, Inc., 2013.

\bibitem{koza1992genetic}
John~R Koza.
\newblock {\em Genetic programming: on the programming of computers by means of
  natural selection}, volume~1.
\newblock MIT press, 1992.

\bibitem{koza2010human}
John~R Koza.
\newblock Human-competitive results produced by genetic programming.
\newblock {\em Genetic Programming and Evolvable Machines}, 11(3-4):251--284,
  2010.

\bibitem{levenberg1944method}
Kenneth Levenberg.
\newblock A method for the solution of certain non-linear problems in least
  squares.
\newblock {\em Quarterly of applied mathematics}, 2(2):164--168, 1944.

\bibitem{salustowicz1997probabilistic}
Rafal Salustowicz and J{\"u}rgen Schmidhuber.
\newblock Probabilistic incremental program evolution.
\newblock {\em Evolutionary Computation}, 5(2):123--141, 1997.

\bibitem{Sthephens2016}
Trevor Stephens.
\newblock {Genetic Programming} in {Python}, with a scikit-learn inspired
  {API}: {gplearn}, 2016--.
\newblock [Online; accessed 21.6.2017].

\bibitem{uy2011semantically}
Nguyen~Quang Uy, Nguyen~Xuan Hoai, Michael O’Neill, Robert~I McKay, and Edgar
  Galv{\'a}n-L{\'o}pez.
\newblock Semantically-based crossover in genetic programming: application to
  real-valued symbolic regression.
\newblock {\em Genetic Programming and Evolvable Machines}, 12(2):91--119,
  2011.

\bibitem{white2013better}
David~R White, James Mcdermott, Mauro Castelli, Luca Manzoni, Brian~W Goldman,
  Gabriel Kronberger, Wojciech Ja{\'s}kowski, Una-May O’Reilly, and Sean
  Luke.
\newblock Better gp benchmarks: community survey results and proposals.
\newblock {\em Genetic Programming and Evolvable Machines}, 14(1):3--29, 2013.

\end{thebibliography}

\end{document}